\documentclass[final,1p,times]{elsarticle}

\usepackage{graphicx}
\usepackage{amssymb}
\usepackage{lineno}

\usepackage{todonotes}
\usepackage{hyperref}
\journal{Elsevier Gait \& Posture:~\url{https://doi.org/10.1016/j.gaitpost.2022.07.153}}

\begin{document}

\begin{frontmatter}

\title{Explaining Machine Learning Models for Age Classification in Human Gait Analysis}

\author[add1]{Djordje Slijepcevic\corref{contrib}}
\author[add2]{Fabian Horst\corref{contrib}}
\author[add2]{Marvin Simak}
\author[add3]{Sebastian Lapuschkin}
\author[add4]{Anna-Maria Raberger}
\author[add3]{Wojciech Samek}
\author[add5]{Christian Breiteneder}
\author[add2]{Wolfgang Immanuel Sch\"ollhorn}
\author[add1]{Matthias Zeppelzauer}
\author[add4]{Brian Horsak}

%\address[add_coauthor]{Both authors contributed equally to this research.}
\address[add1]{Institute of Creative Media Technologies, Department of Media \& Digital Technologies, St. P\"olten University of Applied Sciences, St. P\"olten, Austria}
\address[add2]{Department of Training and Movement Science, Institute of Sport Science, Johannes Gutenberg-University Mainz, Mainz, Germany}
\address[add3]{Department of Video Coding \& Analytics, Fraunhofer Heinrich Hertz Institute, Berlin, Germany}
\address[add4]{Center for Digital Health \& Social Innovation, Department of Health Sciences, St. P\"olten University of Applied Sciences, St. P\"olten, Austria}
\address[add5]{Institute of Visual Computing and Human-Centered Technology, TU Wien, Vienna, Austria}

\cortext[contrib]{Both authors contributed equally to this research.}
% see: https://tex.stackexchange.com/questions/558203/elsarticle-equal-contribution-and-corresponding-author

\end{frontmatter}

\section{Introduction}
\label{S:1}

Machine learning (ML) models have proven effective in classifying gait analysis data~\cite{halilaj2018machine}, e.g., binary classification of young vs. older adults~\cite{begg2005machine,eskofier2013marker,zhou2020detection}. ML models, however, lack in providing human understandable explanations for their predictions. This "black-box" behavior impedes the understanding of which input features the model predictions are based on. We investigated an Explainable Artificial Intelligence method, i.e., Layer-wise Relevance Propagation (LRP)~\cite{bach2015pixel}, for gait analysis data.

\section{Research Question}

Which input features are used by ML models to classify age-related differences in walking patterns?

\section{Methods}

We utilized a subset of the \textsc{AIST Gait Database 2019}~\cite{kobayashi_hida_nakajima_fujimoto_mochimaru} containing five bilateral ground reaction force (GRF) recordings per person during barefoot walking of healthy participants~(Figure 1A). Each input signal was min-max normalized before concatenation and fed into a Convolutional Neural Network (CNN). Participants were divided into three age groups: young (20-39 years), middle-aged (40-64 years), and older (65-79 years) adults. The classification accuracy and relevance scores (derived using LRP) were averaged over a stratified ten-fold cross-validation.

\section{Results}
\label{S:2}

The mean classification accuracy of 60.1$\pm$4.9\% was clearly higher than the zero-rule baseline~(37.3\%). The confusion matrix~(Figure 1B) shows that the CNN distinguished younger and older adults well, but had difficulty modeling the middle-aged adults. 
%A CNN trained for a binary classification of young and older adults, achieved an accuracy of 81.0\%, which is comparable to the classification results of Begg and Kamruzzaman~\cite{begg2005machine}. 
%Figure 1C shows averaged GRF signals with the explainability results.
LRP showed that for young adults, the most relevant regions were the second peak in $GRF_{AP}$ and $GRF_{V}$. For middle-aged adults, regions in $GRF_{ML}$ and the first peak and incline to the second peak of $GRF_{V}$ were most relevant. For older adults, the highest relevance scores were found at the incline to the first peak in $GRF_{V}$ and the end of $GRF_{ML}$.

\begin{figure}[h!]
  \centering	\includegraphics[width=1\linewidth]{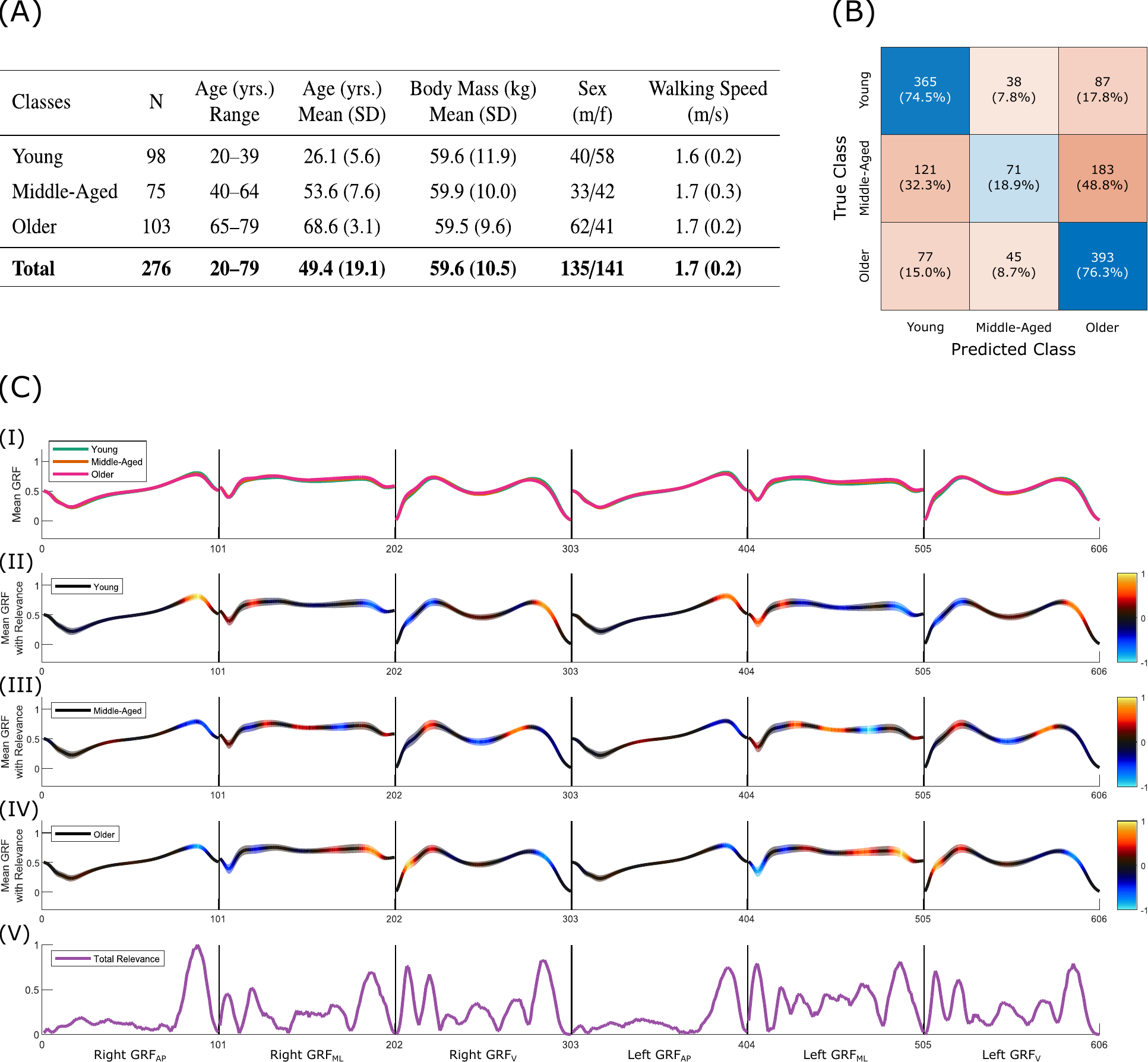} %width=0.90
    \caption{(A) Demographic details, (B) confusion matrix, and (C) explainability results showing (I) averaged GRF signals,
(II)-(IV) averaged GRF signals for each age group with a band of one standard deviation color-coded via LRP relevance scores (yellow in favor and blue against the ground truth class), and (V) total relevance (absolute sum of LRP relevance scores of all classes).}
    \label{img:cnn-norm}
\end{figure}

\section{Discussion}
According to LRP, relevant regions for age classification reside in all GRF signals. Relevant regions at the second peak of $GRF_{AP}$ and $GRF_{V}$ are supported by the literature~\cite{boyer2012role,toda2015age,boyer2017systematic}, whereas the relevant region at the first peak of $GRF_{V}$ is not. Certain relevant regions, e.g., the incline to the first (older adults) and second (middle-aged adults) peak in $GRF_{V}$, as well as regions in $GRF_{ML}$, were not investigated in the past and raise questions for future research.
Our results suggest that gait patterns of middle-aged adults are difficult to model since age-related changes occur at different ages. However, for young and older adults ML models are more effective and ground their predictions on information also identified in the literature~\cite{boyer2017systematic}.

%\section{References}

\section*{Funding}
This work was partly funded by the Research Promotion Agency of Lower Austria (Gesellschaft für Forschungsförderung NÖ) within the Endowed Professorship for Applied Biomechanics and Rehabilitation Research (\#SP19-004) and the FTI Basic Science call (\#FTI17-014). Further support was received from the German Ministry for Education and Research as BIFOLD (\#01IS18025A and \#01IS18037A) and TraMeExCo (\#01IS18056A).

%\begin{table*}[ht!]
%\centering
%\caption{Demographic details of the employed dataset.}
%\label{table:dataset}
%\begin{tabular}{lcccccc}
%\hline
%Classes & N & \begin{tabular}[c]{@{}c@{}}Age (yrs.)\\ Range\end{tabular} & \begin{tabular}[c]{@{}c@{}}Age (yrs.)\\ Mean (SD)\end{tabular} & \begin{tabular}[c]{@{}c@{}}Body Mass (kg)\\ Mean (SD)\end{tabular} & \begin{tabular}[c]{@{}c@{}}Sex\\ (m/f)\end{tabular} & \begin{tabular}[c]{@{}c@{}}Walking Speed\\ (m/s)\end{tabular} \\
%\hline
%Young     & 98  & 20--39  & 26.1 (5.6) & 59.6 (11.9) & 40/58 & 1.32 (0.12) \\
%Middle-Aged   & 75 & 40--64  & 53.6 (7.6) & 59.9 (10.0) & 33/42  & 1.39 (0.20) \\
%Older    & 103 & 65--79  & 68.6 (3.1) & 59.5 (9.6) & 62/41 & 1.35 (0.14) \\
%\hline
%\textbf{Total}		& \textbf{276} & \textbf{20--79} & \textbf{49.4 (19.1)} & \textbf{59.6 (10.5)} & \textbf{135/141} & \textbf{1.35 (0.16)} \\ \hline
%\end{tabular}
%\end{table*}

%% New version of the num-names style
\bibliographystyle{elsarticle-num-names}
\bibliography{sample.bib}






% \end{thebibliography}

\end{document}